\pdfoutput=1
\documentclass[aps,pre,twocolumn,superscriptaddress,floatfix,10pt]{revtex4-2}

\usepackage{mathptmx}
\usepackage{graphicx}
\usepackage{amsmath,amssymb}
\usepackage{booktabs}
\usepackage{xcolor}
\usepackage[colorlinks=true,linkcolor=blue!50!black,citecolor=blue!50!black,urlcolor=blue!50!black]{hyperref}

\graphicspath{{figures/}}

\newcommand{\Da}{D_{a}}
\newcommand{\Dc}{D_{c}}
\newcommand{\Dj}{D_{j}}
\newcommand{\Dd}{D_{d}}
\newcommand{\Esg}{E_{\mathrm{sg}}}
\newcommand{\Etr}{E_{\mathrm{tr}}}
\newcommand{\dd}{\mathrm{d}}
\newcommand{\sci}[2]{#1\times10^{#2}}
\newcommand{\cell}[2]{\parbox[t]{#1}{\raggedright #2\strut}}

\begin{document}

\title{Structure-preserving fidelity diagnostics for partial differential equations}

\author{Kyriakos Flouris}
\affiliation{MRC Biostatistics Unit, University of Cambridge, Cambridge, United Kingdom}
\affiliation{King's College, University of Cambridge, Cambridge, United Kingdom}

\begin{abstract}
Machine learning surrogates of partial differential equations (PDEs) are commonly
trained on data generated by numerical solvers. Therefore, the training data obey not
the PDE but the modified equation of the numerical scheme, and the test loss, computed
on data from the same solver, cannot distinguish a model that learned the physics from
one that learned the discretization error. Here, we show that the difference is
nevertheless visible in the trained model. We propose a suite of structure-preserving
fidelity diagnostics, simple scores that measure whether a learned model respects one
known property of the physics, such as a conserved quantity, a symmetry or a
Hamiltonian structure, without access to the training data or the governing equation.
Over five systems, the diagnostics track the error with respect to the true dynamics
with a rank correlation of up to $0.94$, while the test loss is uncorrelated or
anti-correlated with it. For the Burgers, Korteweg--de Vries and Navier--Stokes
equations, the diagnostics separate numerical schemes that the test loss ranks as
equivalent, including schemes of the same order.
\end{abstract}

\maketitle

\section{Introduction}

Machine learning models of physical systems described by partial differential equations
(PDEs) have become a standard tool, either as fast surrogates of the
solution~\cite{raissi2019,li2021fno,kochkov2021} or as a means to discover the governing
equation from data~\cite{brunton2016,rudy2017}. Since experimental data are scarce,
such models are in most cases trained on data generated by numerical solvers. A
numerical solver, however, only approximates the PDE. In fact, a consistent scheme of
order $q$ on a grid of spacing $h$ solves exactly a slightly different equation, the
modified equation of the scheme~\cite{warming1974,hairer2006}, which differs from the
PDE by terms of order $h^q$. For example, first-order upwinding of the Burgers equation
adds an artificial viscosity proportional to $h$. Consequently, a model that fits the
training data well learns the modified equation rather than the PDE. For time
integrators of ordinary differential equations this effect has been observed and
corrected for when the scheme is known~\cite{zhu2020imde,keller2019}.

In practice, however, the scheme is often not known to the user of a trained model, and
the standard validation of a machine learning model, the test loss, cannot reveal it:
the test data are generated by the same solver as the training data, so the test loss
measures how well the model reproduces the solver, not the physics. As we show in
Section~\ref{sec:burgers}, three models of the Burgers equation trained on data from
three different solvers have test losses within a factor of $1.7$ of each other, while
their errors with respect to the true dynamics differ by a factor of $71$. The test loss
ranks a model that learned an artificial viscosity as good as the model that learned
the physics. Alternatively, checks that compare the model with the governing equation
require the equation and all its coefficients, which are not available in most
applications of learned surrogates.

We postulate that the information is nevertheless stored in the trained model.
Numerical schemes typically break some structural property of the physics, such as a
conserved quantity, a symmetry or a Hamiltonian structure, at the order of their
discretization error. A model that learned the modified equation inherits this
violation, while a model that learned the physics does not. To this end, we propose
fidelity diagnostics, simple scores $D[f_\theta]$ of a trained model $f_\theta$ that are
zero if the model respects a given structural property and positive otherwise. The
diagnostics require only the name of the property, not the governing equation, and are
evaluated on the trained model alone, without the training data and in most cases
without a rollout. We collect them in a diagnostic suite, Table~\ref{tab:diag}, which
lists for each diagnostic what it requires and what it detects, and which can be applied
to other learned models in the same way.

We test the diagnostics on the viscous Burgers equation, the Korteweg--de Vries (KdV)
equation and two-dimensional (2D) Navier--Stokes flow, each discretized by several
schemes that generate the training data of otherwise identical models. We show that the
test loss fails to indicate the error of the models while the diagnostics track it, that
the learned models reproduce the discretization error of their solver quantitatively,
and that different diagnostics detect different schemes. Furthermore, we show which
diagnostics cannot work: linear conserved quantities and self-consistency checks of
learned time-steppers, including the recently proposed semigroup
consistency~\cite{shikhman2026}. Finally, we report a case in which the
diagnostics do not indicate the rollout error.

First, we discuss related work in Section~\ref{sec:related}. We introduce the modified
equation, the learned models and the diagnostics in Section~\ref{sec:method}. In
Section~\ref{sec:results} the experiments and results are presented. The paper finishes with a discussion of the limitations and a summary. The
appendix contains the derivations, the details of the experimental setup and additional
results.

\section{Related work and motivation}
\label{sec:related}

\emph{Learned surrogates.} Neural operators and related
architectures~\cite{li2021fno,lu2021deeponet,brandstetter2022} are trained and evaluated
on data from numerical solvers, as are the standard benchmarks~\cite{takamoto2022}, and
comparisons are often made against weak numerical baselines~\cite{mcgreivy2024}. Here we
address a complementary issue: even a perfect fit to solver data reproduces the solver,
and an evaluation on data from the same solver cannot detect it.

\emph{Physics-informed learning.} Physical knowledge is often imposed during
training, as a penalty on the PDE residual~\cite{raissi2019,karniadakis2021} or through
the architecture~\cite{greydanus2019}, which requires the equation or fixes the structure
by construction. In contrast, we test structural properties after training.

\emph{Modified equations.} The modified equation is a standard tool for
the analysis of numerical schemes~\cite{warming1974,hairer2006}. For ordinary
differential equations learned from time series, it has been shown that the learned
vector field approximates the modified equation of the integration scheme used in
training, and that this can be corrected when the scheme is
known~\cite{zhu2020imde,keller2019}. Similarly, sparse regression methods for equation
discovery~\cite{brunton2016,rudy2017} recover the modified equation when the data are
simulated. Our setting differs in two respects: the scheme is not known, and the
question is asked of the trained model alone.

\emph{Self-consistency.} Learned time-steppers can be checked without labels by
testing whether they compose correctly over time, used as a training
constraint~\cite{chen2023deeposg} and recently as a diagnostic for learned
simulators~\cite{shikhman2026}. We show in Section~\ref{sec:sg} that such checks cannot
detect a discretization error.

\section{Background and method}
\label{sec:method}

\subsection{Modified equation}

Consider a PDE $u_t=\mathcal{N}[u]$ and a numerical scheme with grid spacing $h$. Taylor
expansion of the discrete operator gives the modified equation~\cite{warming1974}
\begin{equation}
u_t=\mathcal{N}_h[u]=\mathcal{N}[u]+h^{q}R[u]+O(h^{q+1}),
\label{eq:modified}
\end{equation}
where $q$ is the order of the scheme and $R$ is the leading error term. For example,
for the Burgers equation $u_t+uu_x=\nu u_{xx}$ with second-order central differences,
$R=-\tfrac16uu_{xxx}+\tfrac{\nu}{12}u_{xxxx}$, a dispersive error. A solution computed
with the scheme satisfies Eq.~\eqref{eq:modified} more accurately than the PDE. Hence,
training data generated by the scheme are samples of $\mathcal{N}_h$, and a learned
model $f_\theta$ that minimizes the training loss approximates $\mathcal{N}_h$. The same
holds for time integrators, where the step size $\Delta t$ takes the role of
$h$~\cite{hairer2006}.

\subsection{Learned models}

Each model $f_\theta$ maps a state $u$ to its time derivative $u_t$ and is trained by
minimizing the mean squared error on pairs $(u,u_t)$, with $u_t$ estimated from
snapshots by a fourth-order finite difference in time. Every scheme is observed
identically: the same snapshot interval, number of samples and observation noise. The
models are stencils with polynomial terms in $u$, linear in their parameters and fitted
in closed form, so that no optimizer variance enters the comparison. Each model class
contains the modified equation of every compared scheme; schemes outside the class are
reported but excluded from the claims (Appendix~\ref{app:setup}).

\subsection{Fidelity diagnostics}

A fidelity diagnostic measures the violation of one structural property $P$ of the
physics by the trained model,
\begin{equation}
D_P[f_\theta]=\frac{\big\langle\|r_P(f_\theta,u)\|^2\big\rangle}
{\big\langle\|s_P(f_\theta,u)\|^2\big\rangle},
\label{eq:D}
\end{equation}
where $r_P$ is a residual that vanishes if $P$ holds, $s_P$ a normalization of the same
size as the terms entering $r_P$, and $\langle\cdot\rangle$ the average over probe
states $u$ that are drawn independently of the training data.

The diagnostic suite is summarized in Table~\ref{tab:diag}. It consists of three
families. The conservation diagnostic $\Da[Q]$ measures whether the model changes a
quantity $Q$ at the rate prescribed by the physics, for example zero for a conserved
momentum or $-\nu\int u_x^2\,\dd x$ for the energy of a viscous flow; the drift
diagnostic $D_{\mathrm{drift}}[Q]$ measures the same along a rollout of the model. The
symmetry diagnostic $\Dc[g]$ measures whether the model commutes with a symmetry $g$ of
the physics, for example a Galilean boost or time reversal. The structure diagnostic
$\Dj$ measures whether the Jacobian of the model has the form required by the physics,
for example the symmetry implied by a Hamiltonian structure, and the recovery
diagnostic $\Dd$ fits a sparse equation to the model's own output and measures the
terms that do not belong to the physics. Each diagnostic requires only the property it
tests, and all except $D_{\mathrm{drift}}$ are evaluated without a rollout. If a scheme
violates the property at order $h^q$, then $D_P[\mathcal{N}_h]=O(h^{2q})$ while
$D_P[\mathcal{N}]=0$. Whether a learned model inherits this value is the question we
test.

\begin{table*}[t]
\caption{The diagnostic suite. Each diagnostic is computed from the trained model alone,
Eq.~\eqref{eq:D}, and requires only the listed property of the physics. Last column:
what the diagnostic detects in this work, with its spread (largest over smallest value)
across the learned models of each system. The last two rows are diagnostics that cannot
detect a solver (Section~\ref{sec:rules}).}
\label{tab:diag}
\begin{ruledtabular}
\begin{tabular}{lllll}
\cell{2.3cm}{diagnostic} & \cell{4.1cm}{tests whether the model} & \cell{2.4cm}{requires} & \cell{0.95cm}{rollout} & \cell{6.0cm}{detects, spread across learned models} \\[3pt]
\colrule
\multicolumn{5}{l}{\emph{Conservation}}\\
\cell{2.3cm}{$\Da[Q]$} & \cell{4.1cm}{changes a quantity $Q$ at the rate of the physics} & \cell{2.4cm}{$Q$ and its rate} & \cell{0.95cm}{no} & \cell{6.0cm}{artificial viscosity and dispersion, Burgers $\sci{9.0}{7}$; non-conservative forms, KdV $123$; energy--enstrophy pair, Navier--Stokes $5.6$; Euler schemes, oscillators $\sci{4.5}{3}$} \\[3pt]
\cell{2.3cm}{$D_{\mathrm{drift}}[Q]$} & \cell{4.1cm}{keeps $Q$ along its own rollout} & \cell{2.4cm}{$Q$} & \cell{0.95cm}{yes} & \cell{6.0cm}{non-symplectic schemes, oscillators $\sci{9.1}{3}$} \\[3pt]
\multicolumn{5}{l}{\emph{Symmetry}}\\
\cell{2.3cm}{$\Dc[g]$} & \cell{4.1cm}{commutes with a symmetry $g$} & \cell{2.4cm}{the symmetry $g$} & \cell{0.95cm}{no} & \cell{6.0cm}{dispersive error, Burgers $\sci{6.0}{3}$ (blind to viscosity); time reversal, oscillators $\sci{1.4}{3}$} \\[3pt]
\multicolumn{5}{l}{\emph{Structure}}\\
\cell{2.3cm}{$\Dj$} & \cell{4.1cm}{has a Jacobian of the required form, e.g.\ Hamiltonian} & \cell{2.4cm}{the structure} & \cell{0.95cm}{no} & \cell{6.0cm}{non-symplectic schemes, oscillators $\sci{8.4}{2}$--$\sci{3.4}{3}$; KdV $3.9$ (limited by the fit)} \\[3pt]
\cell{2.3cm}{$\Dd$} & \cell{4.1cm}{contains no terms outside a sparse equation} & \cell{2.4cm}{candidate terms} & \cell{0.95cm}{no} & \cell{6.0cm}{Burgers $\sci{8.6}{2}$; spurious terms exactly zero except for Euler schemes, oscillators} \\[3pt]
\multicolumn{5}{l}{\emph{Cannot detect a solver}}\\
\cell{2.3cm}{$\Da[c^{\top}u]$} & \cell{4.1cm}{conserves a linear quantity (mass, circulation)} & \cell{2.4cm}{$Q$} & \cell{0.95cm}{no} & \cell{6.0cm}{negative control, silent: spread $\le4.5$} \\[3pt]
\cell{2.3cm}{$\Esg$ and other self-consistency checks} & \cell{4.1cm}{composes consistently in time} & \cell{2.4cm}{nothing} & \cell{0.95cm}{no} & \cell{6.0cm}{detects invalid time-steppers, not solvers: spread $1.04$--$1.08$ across solvers} \\[3pt]
\end{tabular}
\end{ruledtabular}
\end{table*}

To evaluate the diagnostics, every model is also scored by its transfer error $\Etr$,
the root-mean-square relative deviation of a rollout of the model from an accurate
reference solution of the PDE over a time horizon $T$. The transfer error is available
here only because the data are synthetic. It is the quantity a diagnostic should
indicate in practice, where the true solution is unknown.

\subsection{Limits of detection}
\label{sec:rules}

Not every structural property is broken by a numerical scheme. Two simple observations
determine in advance which diagnostics can respond; derivations are given in
Appendix~\ref{app:proofs}.

\emph{Linear conserved quantities.} If a quantity that is linear
in the state, such as the total mass $\int u\,\dd x$, is conserved by the PDE, it is
conserved exactly by every Runge--Kutta time integrator and by every spatial scheme in
flux form. Consequently, the corresponding diagnostic is zero for every scheme and
cannot separate them. We use such diagnostics as negative controls, which must not
separate the models. Quadratic quantities, such as energy, are in general not
preserved~\cite{hairer2006}, and their diagnostics can respond.

\emph{Self-consistency.} A learned time-stepper
$G_\theta(u,\tau)$, which advances a state by a time $\tau$, can be checked for
self-consistency without any knowledge of the physics. For instance, the semigroup
error~\cite{shikhman2026}
\begin{equation}
\Esg=\frac{\|G_\theta(G_\theta(u,s),t)-G_\theta(u,s+t)\|}{\|G_\theta(u,s+t)\|}
\label{eq:esg}
\end{equation}
measures whether two consecutive steps agree with a single step of the combined length.
However, the one-step map of a numerical scheme is, to all orders in $h$, the exact
time-$h$ map of its modified equation~\cite{hairer2006}. A model that learned the
modified equation therefore learned a valid time evolution, only of a different
equation, and satisfies every self-consistency check as well as a model that learned
the physics. Self-consistency checks thus test the learner, not the physics.

\section{Experiments and results}
\label{sec:results}

\subsection{Test loss and diagnostics}
\label{sec:loss}

We first compare the test loss and the diagnostics over all systems, before discussing
each system in turn. For every system, models are trained on data from several schemes
of the same equation, and each model is scored by its test loss, its transfer error and
the largest diagnostic of the suite. In addition to the three PDEs, we include the two
Hamiltonian oscillators of Appendix~\ref{app:odes}, where the error comes from the time
integrator.

\begin{figure*}[t]
\centering
\includegraphics[width=\textwidth]{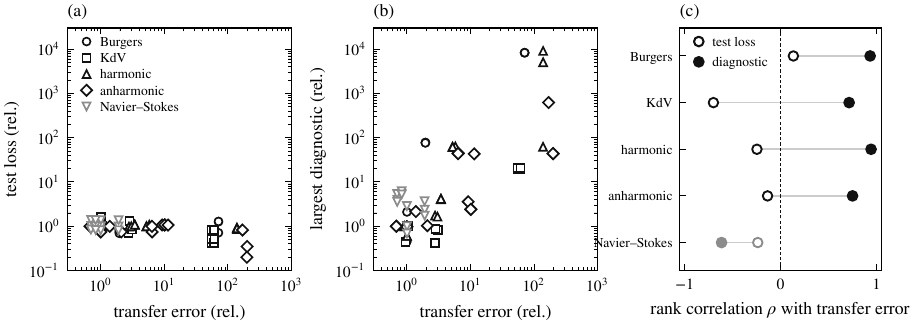}
\caption{Test loss against fidelity diagnostics. Each marker is one learned model
(scheme and seed); both axes are relative to the model trained on the reference scheme
($J_A$ for Navier--Stokes). (a)~Test loss and (b)~the largest diagnostic of the suite
against the transfer error. (c)~Spearman rank correlation with the transfer error per
system, for the test loss (open) and the diagnostic (filled). Navier--Stokes (grey) is
discussed in Section~\ref{sec:ns}.}
\label{fig:loss}
\end{figure*}

Figure~\ref{fig:loss}(a) shows that the test loss is nearly the same for all models of a
system, within a factor of $2.9$, while their transfer errors span more than two orders
of magnitude. The rank correlation of the test loss with the transfer error lies between
$-0.70$ and $+0.13$ [Fig.~\ref{fig:loss}(c)]. It is even negative on four of the five
systems, which we attribute to the reference scheme being the hardest one for a local
model to fit (Appendix~\ref{app:odes}). Consequently, the test loss does not indicate which models
learned the physics. In contrast, the largest diagnostic grows with the transfer error
[Fig.~\ref{fig:loss}(b)], with a rank correlation between $0.71$ and $0.94$ on Burgers,
KdV and the two oscillators. The diagnostics thus expose the models that learned a solver
artifact, while the test loss ranks them as equivalent. The exception is Navier--Stokes,
where the diagnostic is anti-correlated with the transfer error ($\rho=-0.62$;
Section~\ref{sec:ns}).
Across schemes, the strongest diagnostic varies by factors between $5.6$
(Navier--Stokes) and $\sci{9}{7}$ (Burgers), compared with at most $2.9$ for the test
loss. 

\subsection{Burgers equation}
\label{sec:burgers}

We consider the viscous Burgers equation $u_t+uu_x=\nu u_{xx}$ on a periodic domain with
$\nu=0.04$, $n=192$ grid points and smooth random initial conditions. The training data
are generated by four schemes: a pseudo-spectral reference, which stands in for the
physics; second-order central differences; central differences with an added artificial
viscosity $\tfrac12h\,u_{xx}$, a standard stabilization in shock-capturing schemes; and
first-order upwinding, which adds an effective viscosity $\tfrac12|u|h$. The model is a
stencil of radius 5, quadratic in $u$. It contains the first three schemes exactly. The
upwind scheme switches with the sign of $u$ and is outside the model class; it is shown
for completeness only.

The test losses of the three representable models are $\sci{2.4}{-8}$, $\sci{1.8}{-8}$
and $\sci{3.0}{-8}$, while their transfer errors are $\sci{2.4}{-4}$, $\sci{4.6}{-4}$ and
$\sci{1.7}{-2}$: the model trained with artificial viscosity fits its data as well as the
reference model and predicts the true dynamics $71$ times worse.

\begin{figure}[t]
\centering
\includegraphics[width=\columnwidth]{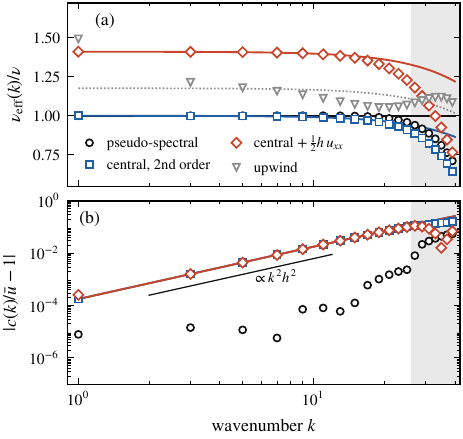}
\caption{Linearized learned Burgers models. (a)~Effective viscosity at $\bar u=0$;
(b)~relative error of the phase speed at $\bar u=0.6$. Markers: learned models trained
on data from each scheme (median over three seeds). Lines: analytic values of the
scheme; dashed: the PDE. The upwind line (grey, dotted) is the averaged prediction
$\nu+\tfrac12\langle|u|\rangle h$. Shaded: wavenumbers at which the training data carry
less than $10^{-6}$ of the peak power, so the model is not constrained by the data.}
\label{fig:burgers}
\end{figure}

The trained model shows why. Linearizing $f_\theta$ around a uniform state $\bar u$
gives, for each wavenumber $k$, an effective viscosity $\nu_{\mathrm{eff}}(k)$ and a
phase speed $c(k)$. For the PDE, $\nu_{\mathrm{eff}}=\nu$ and $c=\bar u$. For central
differences, $\nu_{\mathrm{eff}}=\nu(2-2\cos kh)/(kh)^2$ and $c=\bar u\sin(kh)/(kh)$, and
the artificial viscosity replaces $\nu$ by $\nu+h/2$ (Appendix~\ref{app:schemes}).
Figure~\ref{fig:burgers} compares these analytic values with the learned models. Each
learned model follows the scheme that generated its data, not the PDE. For $k\le16$ the
learned effective viscosity deviates from that of its scheme by at most $1.5\%$, while
the artificial viscosity raises it by $41\%$ above the physical value; the learned
phase speed deviates from that of its scheme by at most $0.04\%$, while the scheme
deviates from the PDE by up to $4.5\%$. Reading the excess viscosity directly from the
model gives $\nu_{\mathrm{eff}}-\nu=0.01631$, compared with the analytic value
$h/2=0.01636$, a ratio of $0.997$ for each of the three training seeds. The model has
learned the modified equation of its solver to within the accuracy of the fit. For the
upwind scheme, which the model can only represent on average, the ratio to the averaged
prediction $\tfrac12\langle|u|\rangle h$ varies between $0.77$ and $1.45$ over seeds.

The diagnostics inherit this difference. The energy diagnostic $\Da$, which uses the
energy decay rate of the PDE and thus requires $\nu$, separates the three representable
models by a factor of $\sci{9.0}{7}$. The Galilean symmetry diagnostic $\Dc$, which
requires only the transformation $u\mapsto u+c$, $x\mapsto x-ct$, separates them by
$\sci{6.0}{3}$. However, $\Dc$ cannot distinguish the central scheme from its
artificially viscous variant ($\sci{3.33}{-6}$ and $\sci{3.34}{-6}$), since a viscosity
term is Galilean invariant while the dispersive error is not. Different diagnostics
detect different schemes, so a single diagnostic is not sufficient.

\subsection{Korteweg--de Vries equation}
\label{sec:kdv}

A diagnostic that only measured accuracy would rank schemes by their order. To test
whether the diagnostics measure structure instead, we use the KdV equation
$u_t=-(3u^2)_x-\delta^2u_{xxx}$ with $\delta^2=0.05$ on $N=127$ grid
points~\cite{zabusky1965}. KdV conserves mass $M=\int u\,\dd x$, momentum
$P=\int u^2\,\dd x$ and the energy $H=\int(u^3-\tfrac12\delta^2u_x^2)\,\dd x$, and it is
Hamiltonian, which implies a symmetry condition on the Jacobian of the right-hand side
that $\Dj$ measures directly (Appendix~\ref{app:schemes}). The nonlinear term
$(3u^2)_x$ can be discretized with the same second-order central stencil in three
forms: the flux form $3D(u^2)$, the advective form $6u\,Du$, and the skew-symmetric
combination of $2/3$ flux and $1/3$ advective form. All three conserve mass, but only
the skew-symmetric form conserves momentum (Appendix~\ref{app:schemes}). We compare
these three forms, a fourth-order skew-symmetric form and a pseudo-spectral reference.

\begin{table*}[t]
\caption{KdV, $N=127$. Diagnostics of the numerical operators for the structure
($\Dj$), mass ($M$), momentum ($P$) and energy ($H$), averaged over 16 probe states.
Values below $10^{-30}$ are round-off. Last column: momentum diagnostic of the model
trained on data from each scheme (median over three seeds).}
\label{tab:kdv}
\begin{ruledtabular}
\begin{tabular}{lccccc}
scheme & $\Dj$ & $\Da[M]$ & $\Da[P]$ & $\Da[H]$ & $\Da[P]$ of the learned model \\
\colrule
spectral       & $<\!10^{-30}$ & $<\!10^{-30}$ & $<\!10^{-30}$ & $<\!10^{-30}$ & $\sci{7.7}{-9}$ \\
skew, 4th      & $\sci{8.8}{-12}$ & $<\!10^{-30}$ & $<\!10^{-30}$ & $\sci{7.9}{-13}$ & $\sci{6.6}{-9}$ \\
skew, 2nd      & $\sci{2.6}{-11}$ & $<\!10^{-30}$ & $<\!10^{-30}$ & $\sci{2.7}{-8}$ & $\sci{1.8}{-9}$ \\
flux, 2nd      & $\sci{1.4}{-9}$  & $<\!10^{-30}$ & $\sci{5.6}{-8}$ & $\sci{2.0}{-9}$ & $\sci{5.6}{-8}$ \\
advective, 2nd & $\sci{7.3}{-9}$  & $<\!10^{-30}$ & $\sci{2.2}{-7}$ & $\sci{2.1}{-7}$ & $\sci{2.2}{-7}$ \\
\end{tabular}
\end{ruledtabular}
\end{table*}

Table~\ref{tab:kdv} shows the diagnostics of the numerical operators. Mass is
conserved by every scheme, as expected from Section~\ref{sec:rules}, so $\Da[M]$ serves
as a negative control. Momentum separates the three second-order forms completely,
from round-off for the skew-symmetric form to $\sci{5.6}{-8}$ and $\sci{2.2}{-7}$ for
the flux and advective forms. The structure diagnostic $\Dj$ is at round-off for the
pseudo-spectral reference and between $10^{-12}$ and $10^{-8}$ for the
finite-difference schemes.

We train a quadratic stencil model of radius 5 on data from each scheme. The test
losses of the five models are within a factor of $2.0$ of each other, while their
transfer errors differ by a factor of $61$. The momentum diagnostic separates the
learned models by a factor of $123$, with the skew-symmetric form lowest (last column
of Table~\ref{tab:kdv}), and the energy diagnostic by a factor of $77$. Thus, the
diagnostics separate schemes of the same order, which differ only in structure, by two
orders of magnitude.

Notably, $\Dj$ also revealed an error in our own reference solver. On a grid with an
even number of points, the highest Fourier mode $k=N/2$ is ambiguous, the quadratic term
aliases into it, and the pseudo-spectral derivative is no longer exactly
antisymmetric. $\Dj$ of the reference operator is $\sci{8.1}{-8}$ at $N=64$, compared
with round-off at $N=63$. Therefore, we use an odd number of grid points throughout.

\subsection{Two-dimensional Navier--Stokes flow}
\label{sec:ns}

We consider 2D incompressible flow in vorticity form,
$\omega_t+J(\psi,\omega)=\nu\nabla^2\omega$ with $\nabla^2\psi=-\omega$, on a periodic
$48\times48$ grid with $\nu=0.05$. The nonlinear term $J$ has three second-order
central discretizations, $J_1$ (advective form), $J_2$ (flux form in $\omega$) and $J_3$
(flux form in $\psi$), which agree in the continuum but not on the grid. Their average
$J_A=\tfrac13(J_1+J_2+J_3)$ conserves both energy $\mathcal{E}$ and enstrophy
$\mathcal{Z}$~\cite{arakawa1966}. The Poisson solve and the viscous term are spectral for
every scheme, so the nonlinear term is the only difference. The model is a $3\times3$
stencil with terms quadratic in $(\omega,\psi)$, which contains all four schemes.

\begin{figure*}[t]
\centering
\includegraphics[width=\textwidth]{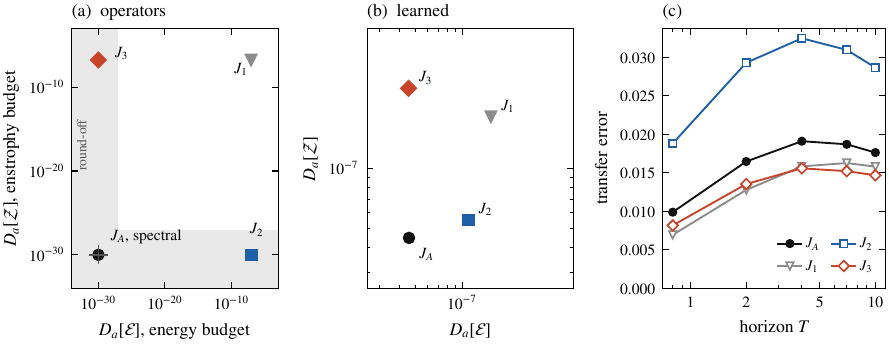}
\caption{2D Navier--Stokes, $48\times48$, $\nu=0.05$. (a)~Energy and enstrophy
diagnostics of the four second-order schemes and the spectral reference; values below
$10^{-27}$ (shaded) are round-off and shown at $10^{-30}$. (b)~The same diagnostics for
models trained on data from each scheme (median over three seeds). (c)~Transfer error of
the learned models against the rollout horizon $T$. The model with the correct
structure ($J_A$, black) has the third-lowest error at every horizon.}
\label{fig:ns}
\end{figure*}

Figure~\ref{fig:ns}(a) shows the energy and enstrophy diagnostics of the four schemes.
$J_2$ conserves only enstrophy, $J_3$ only energy, $J_A$ both and $J_1$ neither, in
agreement with the classical result~\cite{arakawa1966}. Importantly, neither
diagnostic alone identifies the correct scheme: the energy diagnostic cannot
distinguish $J_A$ from $J_3$, and the enstrophy diagnostic cannot distinguish it from
$J_2$. Only the combination isolates it.

Figure~\ref{fig:ns}(b) repeats the measurement on the learned models. The test losses
of the four models are equal ($1.0\times$ spread). The pattern of panel~(a) is preserved,
but the separation shrinks from round-off against $10^{-7}$ to a factor of $5.6$ for
enstrophy and $3.3$ for energy.

On Navier--Stokes, unlike Burgers and KdV, the ranking of the diagnostics does not agree
with that of the transfer error. The $J_A$ model has the lowest diagnostics but the
third-lowest transfer error, at every horizon up to $T=10$ [Fig.~\ref{fig:ns}(c)].
Hence, in this viscous flow the diagnostics identify the scheme with the correct
structure, but they do not indicate the rollout error.

\subsection{Self-consistency checks}
\label{sec:sg}

To test the second observation of Section~\ref{sec:rules}, we compare the semigroup
error, Eq.~\eqref{eq:esg}, with the fidelity diagnostics on the same learned
time-steppers. The model is a time-stepper $G_\theta(u,\tau)=u+\sum_{m\ge1}\tau^m
g_m(u)$, which also provides the vector field $g_1$ for the
diagnostics~\cite{chen2023deeposg}. Fitting such a model with a polynomial library in
the full state is only feasible in low dimensions, so this test uses two Hamiltonian
oscillators, a coupled harmonic and an anharmonic one (Appendix~\ref{app:odes}). We
vary two factors independently: the solver that generates the data, for a fixed model
class, and the model class, for fixed data.

\begin{figure}[t]
\centering
\includegraphics[width=\columnwidth]{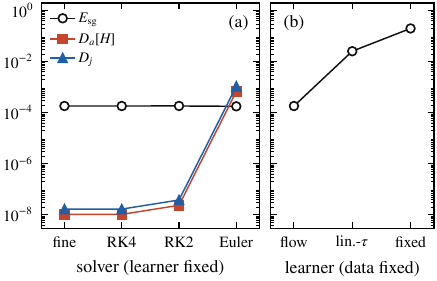}
\caption{Semigroup error and fidelity diagnostics on the same learned time-steppers,
coupled harmonic oscillators, median over three seeds. (a)~Data from four solvers, one
model class: the semigroup error is constant, while the energy diagnostic $\Da[H]$ and
the structure diagnostic $\Dj$ separate explicit Euler by more than four orders of
magnitude. (b)~One data set, three model classes, of which only the first is a valid
time-stepper: the semigroup error separates them.}
\label{fig:sg}
\end{figure}

Figure~\ref{fig:sg} shows the result for the harmonic system. Across solvers, the
semigroup error changes by less than $4\%$, from $\sci{1.79}{-4}$ to $\sci{1.85}{-4}$.
Across model classes, it changes by a factor of $\sci{1.1}{3}$, so the measurement
works and detects models that are not valid time-steppers. On the same models, the
energy and structure diagnostics separate explicit Euler from the reference by
$\sci{6.5}{4}$. The anharmonic system gives the same picture (semigroup error $1.08\times$
across solvers, energy diagnostic $\sci{1.3}{4}$). For a linear system the statement is
exact: the time-stepper that reproduces explicit Euler for a rotation with step $0.25$
corresponds to a vector field that is $12\%$ wrong, and has a semigroup error of
$\sci{4}{-16}$ (Appendix~\ref{app:proofs}). Consequently, self-consistency
checks and fidelity diagnostics are complementary: the former test the learner, the
latter test the physics.

\section{Limitations}

The diagnostics require one structural property of the physics, which must be known. On
learned PDE models they are limited by the fit error of the model: separations of more
than ten orders of magnitude between the numerical operators shrink to factors of $4$ to
$6$ for the KdV structure diagnostic and the Navier--Stokes conservation diagnostics.
Additionally, on Navier--Stokes the diagnostics did not indicate the rollout error. The
models used here are linear in their parameters, which isolates the effect of the solver
from that of the optimization; deep networks, which add optimization error, remain to be
tested. Finally, with five schemes per system the rank correlations
describe orderings; a systematic variation of $h$ at fixed order is needed to measure
the expected $h^{2q}$ scaling directly.

\section{Summary and outlook}

In this work we studied what a learned model of a PDE learns when its training data are
generated by a numerical solver. Over five systems, the test loss, computed on data
from the same solver, is uncorrelated or anti-correlated with the error with respect to
the physics. We showed that the model learns the modified equation of the
solver quantitatively, and that an artificial viscosity is recovered from a learned
Burgers model to within $0.4\%$. We proposed a suite of fidelity diagnostics,
Table~\ref{tab:diag}, which measure whether a trained model respects one structural
property of the physics and require neither the training data nor the governing
equation. The diagnostics track the error on four of the five systems and separate schemes of
the same order for the KdV equation. For Navier--Stokes, they identify the scheme that
conserves both energy and enstrophy, but they do not indicate the rollout error.
Furthermore, we showed that linear conserved quantities and self-consistency checks,
such as the semigroup error, cannot detect a solver.

Notably, different diagnostics detect different schemes, and already for Navier--Stokes
a single property is not sufficient: only the combination of energy and enstrophy
identifies the conservative scheme. More complex equations, for example turbulent or
multi-physics flows, or PDEs whose invariants are not known in closed form, will
therefore require more elaborate diagnostics, combining several properties, resolving
them by scale, or deriving them from the symmetries of the equation. Developing such
diagnostics, and testing them on deep neural surrogates, is the subject of future work.
We anticipate that fidelity diagnostics can serve as a simple validation step for learned
surrogates and discovered equations, in particular when the origin of the training data
is not known.

\bibliography{refs}

\appendix

\section{Schemes and their errors}
\label{app:schemes}

\emph{Burgers.} With the central differences $D_1u_i=(u_{i+1}-u_{i-1})/2h$ and
$D_2u_i=(u_{i+1}-2u_i+u_{i-1})/h^2$, Taylor expansion gives
$D_1u=u_x+\tfrac{h^2}{6}u_{xxx}+O(h^4)$ and
$D_2u=u_{xx}+\tfrac{h^2}{12}u_{xxxx}+O(h^4)$, from which the error term $R$ in
Section~\ref{sec:method} follows. Inserting $u=\bar u+\epsilon e^{ikx}$ into
$-uD_1u+\nu D_2u$ and keeping terms linear in $\epsilon$ gives the growth rate
$-i\bar u\sin(kh)/h-\nu(2-2\cos kh)/h^2$, i.e.\ the phase speed and effective viscosity
quoted in the text. The artificial viscosity replaces $\nu$ by $\nu+h/2$ at every $k$.
Upwinding replaces $D_1$ by a one-sided difference in the direction of
$\mathrm{sign}(u)$, which adds $\tfrac12|u|h\,u_{xx}$ to leading order. A stencil with
fixed coefficients can represent this only on average, $\nu\to\nu+\tfrac12\langle|u|\rangle h$.

\emph{KdV.} Let $D$ be the central difference matrix, which is antisymmetric on a
periodic grid, $a^{\top}Db=-(Da)^{\top}b$. The rate of change of the discrete momentum
$\|u\|^2$ due to the nonlinear term is $-2u^{\top}F(u)$, where $F$ is the discretized
term. With $s=(u^2)^{\top}Du$ one finds $-2u^{\top}\,3D(u^2)=6s$ for the flux form and
$-2u^{\top}(6u\circ Du)=-12s$ for the advective form. A combination with weight
$\alpha$ on the flux form gives $(18\alpha-12)s$, which vanishes for $\alpha=2/3$, the
skew-symmetric form. The dispersive term $D_3u$ is antisymmetric as well and conserves
$\|u\|^2$ for every form. Mass is conserved by every form, because each is a difference
of fluxes, $\mathbf{1}^{\top}D=0$, and $\mathbf{1}^{\top}(u\circ Du)=u^{\top}Du=0$.

The Hamiltonian structure of KdV, $u_t=-\partial_x(\delta H/\delta u)$, implies that
the Jacobian $A$ of the right-hand side satisfies: $\partial_x^{-1}A$ is symmetric on
states with zero mean. $\Dj$ is the relative size of the antisymmetric part of
$\partial_x^{-1}A$, with $\partial_x^{-1}$ applied in Fourier space.

\emph{Navier--Stokes.} The three forms are $J_1=\psi_x\omega_y-\psi_y\omega_x$,
$J_2=(\psi\omega_y)_x-(\psi\omega_x)_y$ and $J_3=(\omega\psi_x)_y-(\omega\psi_y)_x$,
each discretized with central differences on the $3\times3$ block, and
$J_A=\tfrac13(J_1+J_2+J_3)$~\cite{arakawa1966}. The energy and enstrophy production of the nonlinear term alone,
$|\langle\psi,J\rangle|/(\|\psi\|\|J\|)$ and $|\langle\omega,J\rangle|/(\|\omega\|\|J\|)$
(median over 16 probe states), is $\sci{3.0}{-4}$ and $\sci{2.2}{-4}$ for $J_1$,
$\sci{3.2}{-4}$ and $\sci{4.1}{-18}$ for $J_2$, $\sci{3.6}{-18}$ and $\sci{2.2}{-4}$ for
$J_3$, and $\sci{5.6}{-18}$ and $\sci{4.9}{-18}$ for $J_A$.

\section{Derivations}
\label{app:proofs}

\emph{Linear conserved quantities.} One step of a Runge--Kutta method is
$u_{n+1}=u_n+\Delta t\sum_ib_ik_i$, where each stage $k_i$ is an evaluation of the
right-hand side $f$ at some state. If $c^{\top}f(v)=0$ for every state $v$, then
$c^{\top}k_i=0$ for every stage, and $c^{\top}u_{n+1}=c^{\top}u_n$ exactly. Since this
holds for every step size, the modified equation conserves $c^{\top}u$ as well, and so
does a model that learned it. For a quadratic quantity $u^{\top}Cu$ the same argument
fails, and only a special class of implicit methods conserves it~\cite{hairer2006}.

\emph{Self-consistency.} Let $\Phi_h$ be the one-step map of the scheme and $\tilde f$
its modified vector field, so that $\Phi_h$ is the time-$h$ map of $u_t=\tilde f(u)$ to
all orders in $h$~\cite{hairer2006}. A time-stepper trained on data from $\Phi_h$ at
multiples of $h$ approximates the time-$\tau$ map $\phi_\tau$ of $\tilde f$. Every time
evolution satisfies $\phi_s\circ\phi_t=\phi_{s+t}$ and
$\phi_{-\tau}\circ\phi_\tau=\mathrm{id}$, so the semigroup error and the reversibility
defect vanish for $\tilde f$ exactly as for the true field $f$. The remaining errors of
a learned model come from its approximation of $\phi_\tau$, not from the choice of
field. For a linear system $u_t=Au$ with explicit Euler, $\Phi_h=I+hA$ and
$\tilde f(u)=\tilde Au$ with $\tilde A=h^{-1}\log(I+hA)$. For a rotation generator $A$
and $h=0.25$, $\tilde A$ deviates from $A$ by $12.3\%$, while the semigroup error and
the reversibility defect of $\exp(\tau\tilde A)$ are $\sci{3.8}{-16}$ and
$\sci{5.6}{-17}$.

\section{Experimental setup}
\label{app:setup}

\emph{Data.} Snapshots of every scheme, including the reference, are recorded at the
same interval and $u_t$ is estimated from them by a fourth-order central difference in
time. Gaussian noise of relative size
$10^{-5}$ (KdV) and $10^{-4}$ (Navier--Stokes) is added to the snapshots; the noise must
remain below the discretization error, since at $10^{-3}$ the time differencing
amplifies it above the KdV discretization error and all diagnostics read $\sim\!10^{-6}$.
The Burgers data are noise free. The flows are resolved, with a cell Reynolds number
$u_{\max}h/\nu$ of order one; otherwise all schemes are inaccurate and the test loss
separates them for that reason.

\emph{Models.} Burgers: stencil of radius 5, quadratic in $u$, ridge penalty
$10^{-10}$. KdV: quadratic stencil of radius 5; a smaller radius cannot represent the
nonlocal pseudo-spectral reference, which then fits worst (test-loss spread $4736\times$
at radius 3 against $4\times$ at radius 5). Navier--Stokes: $3\times3$ stencil with all
quadratic terms in $(\omega,\psi)$, with $\psi$ obtained from $\omega$ by the spectral
Poisson solve. The pseudo-spectral Navier--Stokes reference is not representable by a
local stencil and is excluded from the comparison of the learned models.

\emph{Diagnostics.} The diagnostics are averaged over 64 (Burgers) or 16 (KdV,
Navier--Stokes) probe states drawn independently of the training data. The
normalization in Eq.~\eqref{eq:D} is $\|\nabla Q\|\,\|f\|$ for $\Da$, the size of the
transformed output for $\Dc$, and the norm of the Jacobian for $\Dj$, which is computed
by finite differences.

\emph{Transfer error.} From eight probe states, the model is integrated with a
fourth-order Runge--Kutta method with step $\le10^{-2}$, and the PDE with the same
method and a $20$ times smaller step. The relative error $\|u_\theta-u\|/\|u\|$ is
averaged in the root-mean-square sense over steps and initial states up to the horizon
$T$. All reported values are medians over three seeds.

\emph{Controls.} Diagnostics that must not separate the schemes, according to
Section~\ref{sec:rules} or by construction, are fixed in advance as negative controls:
KdV mass and translation symmetry, and Navier--Stokes circulation, point inversion and
reflection combined with $\omega\mapsto-\omega$. All controls remain silent, with
spreads across schemes between $1.07$ and $4.5$.

\section{Hamiltonian oscillators}
\label{app:odes}

\emph{Setup.} A coupled harmonic system (frequencies $1.0$ and $1.7$, coupling $0.4$) and
an anharmonic oscillator (quartic coefficient $0.6$). The data are generated by a
fine-step reference ($\Delta t=\sci{2.5}{-4}$) and by explicit Euler, second- and
fourth-order Runge--Kutta and symplectic Euler at $\Delta t=0.05$. The models are
polynomial in the state, with the degree of the true system, and the snapshots carry
observation noise of relative size $10^{-3}$.

\emph{Same-order schemes.} Explicit and symplectic Euler are both first
order. Their test losses are equal (ratio $0.90$ and $0.35$ for the two systems), their
transfer errors differ by factors of $27$ and $31$, and the energy diagnostic differs
only by factors of $1.9$ and $2.9$, since both schemes have an energy error of order
$\Delta t$ per step. The structure diagnostic $\Dj$, which tests the Hamiltonian form of
the Jacobian, separates them by factors of $\sci{8.0}{2}$ and $\sci{3.4}{3}$.

\emph{Time-steppers.} The time-stepper of Section~\ref{sec:sg} is cubic in the state and
in $\tau$, fitted at lags of one to three snapshot intervals without noise; the
alternatives, linear in $\tau$ and a fixed-step map, are not valid time-steppers. The
reversibility defect and the equal-split semigroup error change by at most $1.27\times$
across solvers on both systems.

\emph{Other metrics.} On the two oscillators we compared $31$ metrics by their mean
absolute rank correlation with the transfer error: the energy diagnostic reaches $0.90$, a
metric that requires the full equation with its parameters $0.89$, an ensemble of models
$0.68$, and every metric computed from held-out data at most $0.42$ (test loss $0.19$).
We attribute the negative correlation of the test loss in Fig.~\ref{fig:loss}(c) to the fact that the
reference is the hardest scheme for a local model to fit.

\end{document}